\documentclass{Interspeech2024}

\usepackage{cite}
\usepackage{url}
\usepackage{makecell, hhline}
\usepackage{float}
\usepackage{amsmath,amssymb,amsfonts}
\usepackage{algorithmic}
\usepackage{graphicx}
\usepackage{textcomp}
\usepackage{xcolor}
\usepackage[nolist,nohyperlinks]{acronym}
\usepackage{multirow}
\usepackage[subscriptcorrection,varg,varvw,cmbraces,cmintegrals]{newtxmath}
\usepackage{subcaption}
\usepackage[style=base]{caption}

\interspeechcameraready

\title{Test-Time Training for Depression Detection}
\name[affiliation={1,2}]{Sri Harsha}{Dumpala}
\name[affiliation={1,2}]{Chandramouli Shama}{Sastry}
\name[affiliation={1,3}]{Rudolf}{Uher}
\name[affiliation={1,2}]{Sageev}{Oore}

\address{
  $^1$Dalhousie University, Canada; $^2$Vector Institute, Canada; $^3$Nova Scotia Health, Canada}
\email{\{sriharsha.d, cssastry, uher, sageev\}@dal.ca}

\begin{document}

\maketitle
\begin{abstract}
Previous works on depression detection use datasets collected in similar environments to train and test the models. In practice, however, the train and test distributions cannot be guaranteed to be identical. Distribution shifts can be introduced due to variations such as recording environment (e.g., background noise) and demographics (e.g., gender, age, etc). Such distributional shifts can surprisingly lead to severe performance degradation of the depression detection models. In this paper, we analyze the application of test-time training (TTT) to improve robustness of models trained for depression detection. When compared to regular testing of the models, we find TTT can significantly improve the robustness of the model under a variety of distributional shifts introduced due to: (a) background-noise, (b) gender-bias, and (c) data collection and curation procedure (i.e., train and test samples are from separate datasets). 
\end{abstract}

\section{Introduction}
\label{intro}
Depression is one of the most common mental health disorder and a leading cause of disability worldwide \cite{rehm2019global}. According to the World Health Organization~\cite{WHO_2022}, depressive disorders are highly prevalent worldwide, yet remain largely under-detected and under-treated~\cite{herrman2022time}. Extensive screening and early diagnosis of depressive symptoms is crucial to control the depression severity. Inadequate accessibility to clinical services, that is further accentuated by associated stigma impedes early detection. Automated assessment systems can help in early detection allowing individuals to seek timely professional help. To enable widespread adoption of automated assessment systems, it is important to develop reliable and robust systems for depression detection and is the focus of this work.  

Depression assessment from speech has attracted significant research interest since natural speech contains necessary information and markers for depression assessment and a variety of handheld devices offer recording functionality\cite{cummins2015review, low2020automated, dikaios2023applications}. Following standard machine learning practices, most of the previous research works on depression assessment study audio models that are trained and evaluated on data collected in similar environments i.e., matching train and test conditions. In practice, however, the train and test distributions cannot be guaranteed to be identical, i.e., there may be a distributional shift between the train and test data. Distribution shifts in speech can arise due to: (a) inter-speaker variations such as speaking style, gender, age; (b) recording environment can introduce different background noises such as babble, living room, traffic, etc. Such distributional shifts can surprisingly lead to severe performance degradation even in state-of-the-art deep learning models \cite{ASR_benchmarking, garcia2019speaker, cross_corpus_emotion, botelho2022challenges}. The limited availability of large-scale depression datasets further increases the vulnerability to distribution shifts; on the other hand, large-scale depression datasets that cover \textit{several} distribution shifts are not only expensive and challenging to acquire but can also not guard against distribution shifts not included in training data. Thus, we aim to improve the robustness of depression detection models trained with available (limited) training data and instead utilize test-time training as described below. 


Test-time training (TTT) \cite{sun2020test, liu2021ttt++, ttt_mae} is extensively studied in applications such as image classification and demonstrated to offer improved robustness against a variety of (unseen) distribution shifts. In TTT, a part of the model parameters are updated based on a self-supervised loss objective (i.e., does not require any labels) only using the test-sample. The updated parameters are specific to each test sample and reset after the prediction has been made.  

The efficacy of TTT is determined by the self-supervised learning task \cite{liu2021ttt++} and \cite{ttt_mae} demonstrates that masked auto-encoding offers reliable improvements against distribution shifts.
Motivated by the success of the transformer-based masked autoencoders (MAE) for speech \cite{mae_audio_neurips}, we extend a test-time training approach based on MAE \cite{ttt_mae, dumpala2023ttt} to depression detection in this work. 
To the best of our knowledge, this is the first work to use test-time adaptation for depression detection. We show that TTT-MAE for depression detection achieves significant improvements under a variety of distributional shifts. In particular we experiment with the following types of distributional shift in this work:
1) Background noise: trained with clean speech but tested on speech corrupted with background noise; 2) Gender-bias: trained with speech samples from female speakers and evaluated over speech samples from male speakers and vice-versa; 3) Dataset: train data and test data are obtained from different datasets --- the distribution shift between training samples and testing samples when obtained from different datasets is not easy to define and is rooted in the data collection and curation procedure. 



\section{Related Work}
\noindent\textbf{Depression detection under distributional shifts: }
Most of the previous studies on depression detection utilized speech recordings collected in controlled laboratory settings. Deploying depression screening systems in-the-wild (for instance using smartphones in natural environments) is still a challenge and relatively less explored.
Few studies analyzed depression detection systems when there is a distributional shift between the train and test conditions \cite{mitra2015cross, gerczuk2023noise, alghowinem2016cross, huang2018depression, rutowski2022toward, seneviratne2020generalized}. These studies showed that performance of the depression assessment systems degrade significantly when there exists a distribution shift between the train and test conditions. In \cite{alghowinem2016cross}, support vector machines (SVMs) models trained with data collected in one country and tested with datasets collected in a different country showed significant performance degradation, even though the language (English) was the same. In \cite{huang2018depression}, SVMs trained using data collected from one set of smartphones and tested with another set of smartphones showed significant performance degradation. This study also found that variation in terms of gender and the speech elicitation approach also resulted in degraded performance. Moreover, models trained with speech data collected from general population with ages ranging from 18 to 65 years yield poor detection rates when evaluated over speech data collected from older population with ages ranging from 45 to 75 years -- despite being trained with a large dataset, and utilizing a pre-trained acoustic model (CNN-based) for speech recognition \cite{rutowski2022toward}. Interestingly, even variation in treatment type can result in performance degradation: for example, \cite{seneviratne2020generalized} report that the performance of the dilated CNN models degraded when train and test speech datasets are collected from participants with and without pharmacotherapy and/or psychotherapy treatment. The above works demonstrate the vulnerability of simple depression detection systems such as SVMs, CNNs and dilated CNN networks trained using conventional acoustic features under distributional shifts. 
In this work, we evaluate self-supervised learning (SSL)-based models in terms of their robustness to distributional shifts in the context of depression detection. While the self-supervised training on large unlabelled datasets render these models more robust than conventional acoustic features, we can still notice significant degradation in performance under distributional shifts.

\begin{figure}[tb]
\vspace{-0.3cm}
  \centering
  \includegraphics[width=0.9\linewidth]{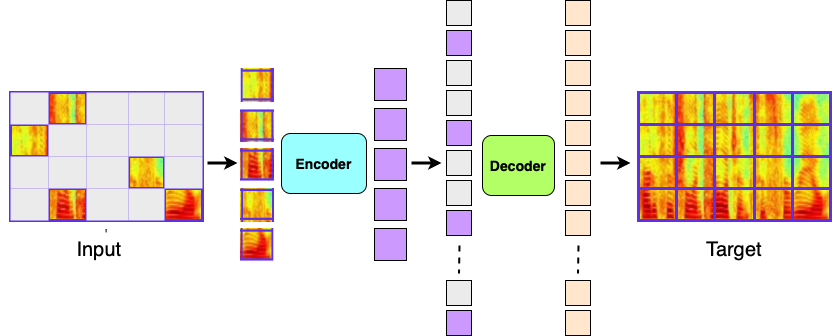}
  \caption{Schematic outline of AudioMAE pre-training.}
  \label{fig:mae_pretrain}
\end{figure}

Robustness to distributional shifts is an unexplored topic in depression detection \cite{huang2020domain}. In \cite{huang2020domain}, domain adaptation techniques were used to improve the performance of depression detection for cross-corpus testing. In domain-adaptation, the model is trained to generalize to all possible distributional shifts. But anticipating every possible distributional shift at train time is not feasible, particularly in real world applications. Most models trained using domain generalization techniques are fixed during inference even when the test distribution changes. 
In this work, we extend TTT techniques for depression detection in presence of distributional shifts.

\noindent\textbf{Pre-trained models for depression detection:} 
Recent studies explored pre-trained models, particularly SSL models for depression detection \cite{zhang2021depa, campbell2023classifying, dumpala2022combining, dumpala2023manifestation, ssl_dep_2023}. But all these studies evaluated SSL-based depression detection under matched train-test conditions. None of the previous work evaluated these models under mismatched train-test conditions. In this paper, we aim to close this gap by analyzing SSL speech models such as Wav2Vec 2.0 \cite{baevski2020wav2vec}, HuBERT \cite{hsu2021hubert}, WavLM \cite{chen2022wavlm} and AudioMAE \cite{mae_audio_neurips} when evaluated with distributionally shifted test instances. 

\noindent\textbf{Test-time training (TTT):} The basic paradigm in TTT ~\cite{sun2020test} is to use a test-time task (usually a self-supervised learning task) besides the main task during  training, and update the pre-trained model using test data with the (self-supervised) test-time objective before the final prediction. In vision, different self-supervised tasks used for TTT include rotation prediction \cite{sun2020test}, contrastive loss \cite{liu2021ttt++} and masked autoencoding \cite{ttt_mae}. Later, TTT-MAE was extended to speech-based tasks such as speaker recognition, emotion classification and short word detection \cite{dumpala2023ttt}. In this work, we extend TTT-MAE framework to depression detection. 

\section{Method}
\label{method}

\subsection{Pre-training MAE.} 
In this paper, we use the audio masked autoencoder (AudioMAE) \cite{mae_audio_neurips}.
AudioMAE is pre-trained to reconstruct the masked patches of speech Mel-spectrogram with an asymmetrical encoder-decoder architecture. In the following, we provide a brief overview of the MAE pre-training (see Figure \ref{fig:mae_pretrain}). 

First, the input speech waveform is transformed into mel-spectrograms, which is then divided into a sequence of non-overlapping grid patches. 
These patches are then flattened and embedded by a linear projection layer. To provide positional information, fixed sinusoidal positional embeddings are added to the embedded patches. Afterward, we randomly mask 80\% of the patches while preserving  positional indices of all the patches. This enables the decoder to reconstruct the spectrogram. For the encoder, only unmasked patches are used to generate latent representations. The decoder then tries to reconstruct the original spectrogram, given  latent representations of the encoder and  masked patches as input. The latent representations and masked patches are organized in the initial order before being provided as input to the decoder.
During training, the objective is to minimize mean squared error (MSE) between reconstructed and input spectrograms, averaged over the masked patches.
We use the pre-trained AudioMAE model released by \cite{mae_audio_neurips} for all the experiments in this paper. 

\begin{figure}[tb]
\vspace{-0.55cm}
  \centering
  \includegraphics[width=0.73\linewidth]{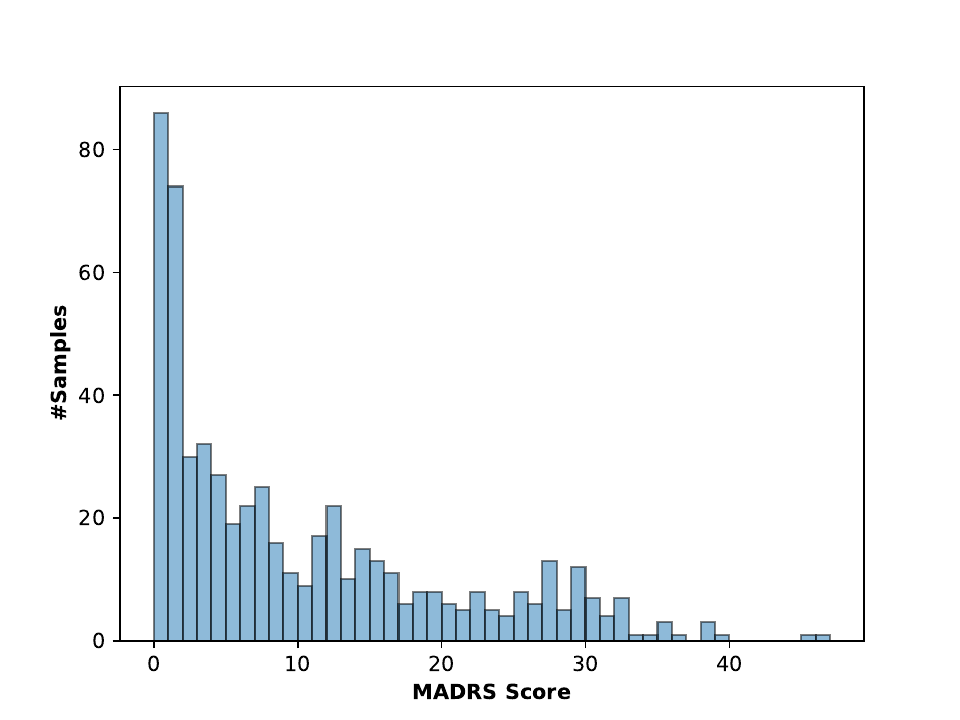}
  \vspace{-0.1cm}
  \caption{\small Distribution of CLD dataset samples with respect to depression severity as measured by MADRS.}
  \label{fig:data_dist}
\end{figure}

\begin{table*}[tb]
\vspace{-0.5cm}
\caption{\small Comparison of AudioMAE-TTT with non-TTT methods (AudioMAE, AudioMAE-FT, other SSL methods (Wav2Vec, HuBERT, WavLM), and conventional speech features (COVAREP, eGeMAPS)) under different different shifts due to background noises (noises added at 5 dB). AudioMAE-TTT significantly outperforms the non-TTT approaches across all the distributional shifts. Each cell contains F-scores in the form: $F_{M}$ ($F_{H}$, $F_{D}$).} 
\label{tab:bg_noises}
\vspace{-0.2cm}
\centering
\resizebox{1.0\linewidth}{!}{%
\begin{tabular}{l|cccccccc|c}
\toprule
Model & \multicolumn{1}{c}{Clean} & \multicolumn{1}{c}{AWGN} & \multicolumn{1}{c}{AC} & \multicolumn{1}{c}{Babble} & \multicolumn{1}{c}{Living Room} & \multicolumn{1}{c}{Park} & \multicolumn{1}{c}{Reverberation} & \multicolumn{1}{c|}{Traffic} & \multicolumn{1}{c}{Average} \\
\midrule
COVAREP & 59.2 (71.8, 46.6) & 35.8 (68.2, 3.6) & 38.7 (56.1, 20.8) & 33.2 (66.4, 0.0) & 40.4 (62.3, 18.1) & 41.9 (68.6, 15.1) & 40.7 (54.5, 26.8) & 49.6 (68.6, 30.7) & 42.4 (64.6, 20.2) \\
eGeMAPS & 63.5 (72.4, 54.7) & 35.1 (61.4, 8.8) & 40.6 (53.6, 27.4) & 33.2 (66.4, 0.0) & 42.1 (61.5, 22.6) & 43.5 (69.4, 17.6) & 44.6 (57.9, 31.3) & 51.4 (66.7, 36.1) & 44.3 (63.7, 24.8) \\
Wav2Vec 2.0 & 67.7 (68.5, 66.8) & 37.2 (56.8, 17.5) & 52.4 (67.3, 36.9) & 35.8 (65.5, 4.1) & 50.7 (68.9, 32.5) & 51.5 (68.1, 34.7) & 48.2 (59.2, 37.2) & 54.3 (70.4, 38.2) & 49.7 (65.7, 33.5) \\
HuBERT & 69.8 (70.3, 69.1) & 40.6 (52.3, 28.6) & 54.1 (65.7, 42.5) & 36.2 (65.1, 7.3) & 52.0 (54.3, 49.6) & 53.0 (60.3, 46.2 & 49.1 (58.5, 39.7) & 55.9 (63.4, 48.4) & 51.4 (61.3, 41.4)  \\
WavLM & 70.9 (73.2, 68.5) & 38.6 (51.1, 26.2) & 57.4 (67.5, 47.1) & 38.3 (63.7, 12.6) & 53.2 (68.4, 38.1) & 55.8 (68.9, 42.6) & 50.7 (63.4, 38.1) & 57.2 (67.3, 47.1) & 52.7 (65.3, 40.1) \\
AudioMAE & 69.4 (70.7, 68.2) & 39.7 (55.6, 24.4) & 55.2 (63.2, 46.9) & 35.9 (63.3, 8.4) & 51.8 (61.3, 42.4) & 53.4 (62.2, 44.6) & 48.8 (60.5, 37.2) & 56.3 (66.4, 46.1) & 51.3 (62.8, 39.7) \\
AudioMAE-FT & 71.1 (73.6, 68.6) & 39.2 (57.1, 21.2) & 53.6 (62.8, 44.4) & 35.5 (64.5, 6.4) & 51.1 (60.7, 41.5) & 52.3 (65.1, 39.7) & 47.5 (59.4, 35.6) & 55.8 (67.1, 44.5) & 50.8 (63.7, 37.8) \\
AudioMAE-TTT & \textbf{71.4 (72.5, 70.3)} & \textbf{52.5 (62.8, 41.7)} & \textbf{62.1 (60.4, 63.7)} & \textbf{47.3 (68.5, 26.3)} & \textbf{58.6 (64.2, 53.1)} & \textbf{60.3 (63.7, 57.2)} & \textbf{59.4 (67.6, 51.3)} & \textbf{63.4 (68.5, 58.3)} & \textbf{59.5 (66.2, 52.7)} \\
\bottomrule
\end{tabular}}
\vspace{-0.2cm}
\end{table*}

\subsection{TTT using Pre-trained AudioMAE}
\textbf{Architecture.} Similar to \cite{sun2020test, ttt_mae}, we use a Y-shaped architecture: a shared encoder network $e$ followed by two heads, a self-supervised head $g$ and a depression detection head $d$. In this work, $e$ and $g$ are the encoder and decoder networks of the pre-trained AudioMAE, respectively and $d$ consists of two feed-forward layers with SiLU activation followed by an output softmax layer for depression detection. We start with the pre-trained AudioMAE.

TTT involves two levels of training 1) train-time training and 2) test-time training

\noindent\textbf{Train-time training:} 
In train-time training, we train the model using labelled data on the downstream task -- here it is depression detection. In train-time training, we only use the encoder and discard the decoder. We further freeze the weights of the encoder (use it as a feature extractor), and only train the depression detection head, where the latent representations generated by the encoder are provided as input to the depression detection head $d$. 


\noindent\textbf{Test-time training.}
At test time, we start from the depression detection head trained above, as well as the AudioMAE pre-trained encoder $f_0$ and decoder $g_0$. For each test input, while freezing the weights of decoder, we optimize only the encoder using the same loss used to pre-train AudioMAE i.e., minimize the MSE between the reconstructed and input masked spectrograms. 
At test-time, all the parameters of the shared encoder are updated to minimize the self-supervised loss across various augmentations of a single test sample.

\section{Dataset Details}
In this work, we use two different speech depression datasets:

\noindent 1) We use an internal depression dataset called Clinically-Labelled Depression dataset (CLDD). CLDD consists of speech samples collected from $559$ ($401$ female and $158$ male) participants. Each participant was prompted with with three different prompts to speak about their experiences from the past few weeks. The three prompts were designed to evoke neutral, positive, and negative context, respectively. Each participant spoke uninterruptedly for at least $3$ minutes for each prompt, resulting in a total of approximately $10$ minutes of speech per speaker. 
Depression severity of each speech sample was scored on the Montgomery and Asberg Depression Rating Scale (MADRS) \cite{madrs_scale}, which is in the range of $0-60$. The range of total MADRS scores in our dataset range from 0 to 47. Participants with MADRS $\geq 10$ are classified as depressed and the remaining are classified as non-depressed (healthy). The distribution of the samples is shown in Figure \ref{fig:data_dist}. The dataset is divided into train and test set with $417$ ($307$ female and $110$ male) and $142$ ($94$ female and $48$ male) recordings, respectively.

We use DAIC-WOZ \cite{gratch2014distress} corpus for cross-data evaluation task. DAIC-WOZ dataset contains a set of $219$ clinical interviews collected from $219$ participants ($154$ healthy and $65$ depressed). Each audio sample was labeled with a PHQ-8 (Patient Health Questionnaire) score, in the range of $0-24$, to denote the severity of depression. For DAIC-WOZ dataset, speech samples with PHQ-8 scores $\geq$ $10$ were considered as depressed and those samples with PHQ-8 scores $<$ $10$ were considered as healthy. We use the train, validation and test splits of DAIC-WOZ as defined in \cite{valstar2016avec, gratch2014distress}

Manual transcripts with timestamps of the DAIC-WOZ and CLD datasets were used to discard the interviewer speech and retain only the participant's speech.
To train and test the models, we segment each speech recording into 7-second segments. The depression label assigned for each segment is same as the label of the entire speech sample. Performance is reported by using majority voting across all segments of test samples.


\section{Experiments}
\subsection{Models for comparison}
In this work along with AudioMAE, we will evaluate the performance of following SSL-based speech models for depression detection under different distributional shifts: (1) Wav2Vec 2.0 \cite{baevski2020wav2vec}, (2) HuBERT \cite{hsu2021hubert} and (3) WavLM \cite{chen2022wavlm}.
We freeze the weights of the pre-trained models and train a one-layer fully connected neural network (with 100 sigmoid linear units (SiLU) \cite{silu_activation}) with an output softmax layer on top of these models for depression detection. 

We also evaluate CNN models trained using conventional speech features which include COVAREP~\cite{degottex2014covarep} and eGeMAPS~\cite{eyben2015egemaps} as baselines. These CNN models comprise two convolutional layers, each with 100 channels. The kernels have sizes of 4 and 5 for the first and second layers, respectively. Outputs from the second convolutional layer are flattened before passing through a fully-connected layer with 100 units and an output layer. 
We extract 88-dimensional eGeMAPS and 74-dimensional COVAREP using OpenSMILE~\cite{eyben2010opensmile} and COVAREP toolkits, respectively.

\subsection{Model training}
To train models (both TTT and non-TTT) on the depression detection task, we freeze all the weights of the pre-trained model and only train the feed-forward neural network and the output layer. We train the feed-forward and output layers using the Adam optimizer with a learning rate of 1e-3, a weight decay of 1e-5, and a batch size of 32. We use Negative log-likelihood (NLL) as the training objective. Each model is trained for 5 epochs.

\noindent\textit{TTT at inference for AudioMAE-TTT}: Following \cite{ttt_mae, dumpala2023ttt}, for each test sample, we train only the encoder (freezing the decoder weights)for $20$ steps with a batch-size of 128 using SGD optimizer with a fixed learning rate of $2.5e-3$, momentum of $0.9$ and weight decay of $0.2$. During TTT, we follow similar procedure as pre-training: mask $80\%$ of the input patches and provide the unmasked patches as input to the encoder whereas all the patches are provided to the decoder. The encoder weights are then updated to optimize the reconstruction loss (MSE) over the masked patches. We note that we do not use any augmentation beyond random masking for TTT. We performed experiments using $4$ Nvidia A40 $48$GB GPUs.

We train CNNs on the conventional speech features using Adam optimizer with a learning rate of $0.001$, weight decay of $1e-5$, and a batch size of $32$. Dropout rates of $0.3$ and $0.4$ were used for the convolutional and fully connected layers, respectively to avoid model overfitting.

A randomly selected subset ($10\%$) of the training set is allocated as validation set for selecting the model hyperparameters.

\begin{table}[tb]
\vspace{-0.3cm}
\caption{\small Distributional shift due to (a) gender variations, (b) dataset variations. Performance (in $F_M$) when models are trained on one gender (dataset) and tested on the other. AudioMAE-TTT significantly outperforms all other methods.} 
\vspace{-0.2cm}
\label{tab:gen_data}
\begin{subtable}{1.0\linewidth}
\caption{\small Distributional shift due to gender variations.}
\label{tab:gender}
\centering
\resizebox{1.0\linewidth}{!}{%
\begin{tabular}{l|cccc}
\toprule
Train set & \multicolumn{2}{c}{Female} & \multicolumn{2}{c}{Male} \\
\midrule
Test set & Female & Male & Male & Female \\
\midrule
Wav2Vec 2.0 & 70.6 & 51.8 & 71.6 & 53.3  \\
HuBERT & 71.8 & 52.3 & 72.3 & 54.2 \\
WavLM & \textbf{73.6}  & 47.3  & \textbf{73.1} & 52.9 \\
AudioMAE & 72.2 & 49.6 & 71.8 & 51.8 \\
AudioMAE-TTT & 73.4 & \textbf{63.1} & 71.8 & \textbf{62.2} \\
\bottomrule
\end{tabular}}
\vspace{0.02cm}
\end{subtable}

\vspace{0.4cm}

\begin{subtable}{1.0\linewidth}
\vspace{-0.1cm}
\caption{\small Distribution shifts due to dataset variation.}
\label{tab:cross_data}
\centering
\resizebox{1.0\linewidth}{!}{%
\begin{tabular}{l|cccc}
\toprule
Train Dataset & \multicolumn{2}{c}{CLD} & \multicolumn{2}{c}{DAIC} \\
\midrule
Test Dataset & CLD & DAIC & DAIC & CLD \\
 \midrule
Wav2Vec 2.0 & 67.7 & 37.9 & 60.4 & 41.2 \\
HuBERT & 69.8 & 40.6 & 64.2 & 43.7 \\
WavLM & 70.9 & 41.5 & \textbf{66.7} & 44.8 \\
AudioMAE & 69.4 & 39.4 & 63.1 & 44.2 \\
AudioMAE-TTT & \textbf{71.4} & \textbf{48.7} & 65.9 & \textbf{56.8} \\
\bottomrule
\end{tabular}}
\end{subtable}
\end{table}

\subsection{Results}
We report the performance of the models in terms of macro F-score ($F_{M}$) which is computed as $F_{M}$ = ($F_1(H)$ + $F_1(D)$)/2, where $F_1(H)$ and $F_1(D)$ are the F-scores of the healthy class and the depressed class, respectively. Unless specified otherwise , we report results for AudioMAE-TTT after $20$ TTT steps.

Table \ref{tab:bg_noises} compares the performance of TTT (AudioMAE-TTT) with no-TTT (Wav2Vec 2.0, HuBERT, WavLM)  under different distributional shifts caused due to background noises. To evaluate models' performance under different distributional shifts, we introduce diverse background noises sourced from Microsoft's Scalable Noisy Speech Dataset (MS-SNSD) \cite{reddy2019scalable}. These noises are exclusively added during the testing phase and are not used in pre-training or train-time training of the self-supervised models.
When trained and tested with clean speech, non-TTT methods such as SSL-based models (Wav2Vec 2.0, HuBERT, WavLM and AudioMAE) and models trained using conventional speech features (COVAREP and EGeMAPS), show significant degradation in performance with SSL-based models performing better than conventional speech models. Whereas, AudioMAE with TTT (AudioMAE-TTT) achieves best performance under all distribution shifts, with lower degradation in performance compared to testing with clean speech. It is interesting to see that Audio-FT (where we finetune the encoder of AudioMAE along with the depression detection head for depression detection task) performs inferior to AudioMAE (where we freeze the weights of the AudioMAE encoder). This is in agreement with the findings in \cite{LPFT2022}.

Table \ref{tab:gen_data} compares the performance of TTT with non-TTT techniques under (a) gender-based and (b) dataset-based distributional shifts. For gender-based experiments, we train the models with all speech samples from the same gender (Female or Male) and test the models on samples other gender (Male of Female). It can be observed from Table \ref{tab:gender} that while non-TTT techniques show significant performance degradation for cross-gender testing compared to same gender testing, AudioMAE-TTT shows relatively low degradation in performance.

For the case of cross-dataset testing, we train the models with one dataset (CLD or DAIC) and test the datasets with another dataset (DAIC or CLD), where CLD is spontaneous speech whereas DAIC is interview-based speech. While non-TTT techniques show significant performance degradation for cross-dataset testing, AudioMAE-TTT outperforms all the other methods with relatively very low degradation in performance.

\begin{figure}[tb]
\vspace{-0.64cm}
  \centering
  \includegraphics[width=0.73\linewidth]{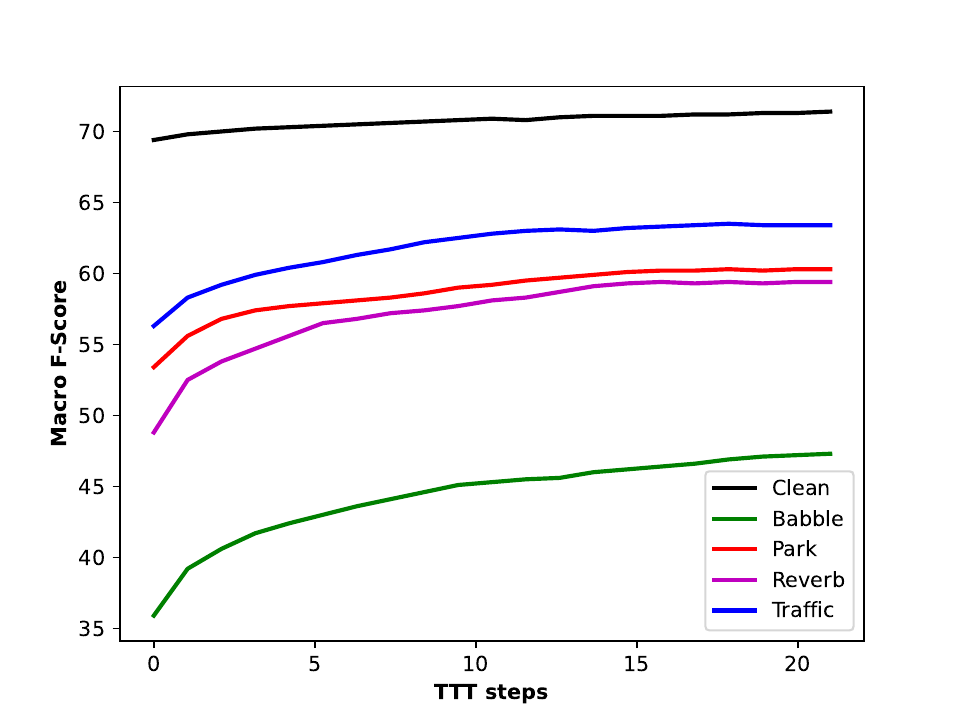}
  \vspace{-0.1cm}
  \caption{\small Performance (in terms of $F_M$) of AudioMAE-TTT across TTT steps. Performance of TTT improves with the number of steps. In this paper, we report results after 20 TTT steps.}
  \label{fig:steps_ttt}
  \vspace{-0.2cm}
\end{figure}

Figure \ref{fig:steps_ttt} shows how the performance of AudioMAE-TTT varies with the number of TTT steps at inference. We show performance curves for $20$ TTT steps, when tested with distributional shifts due to background noises. We can see that the performance of TTT improves as we increase the number of TTT steps. But just a few TTT steps are sufficient to achieve significant improvements in performance. In this paper, we report most of the results after 20 TTT steps.

We evaluate the statistical significance of the improvements obtained by TTT by computing the confidence intervals between TTT and non-TTT approaches \cite{conf_intervals}. For the $95\%$ confidence intervals, we observed no significant overlap between TTT and non-TTT approaches. Furthermore, the confidence interval for the difference in metrics does not contain 0, implying a statistically significant difference in performance between the two types of models.


\section{Conclusions}
Robust and reliable depression detection in the presence of distribution shifts is a challenging problem for both conventional deep neural networks and self-supervised models pre-trained over large datasets. We discuss and evaluate test-time training technique as a solution to achieve robust depression detection even on distributionally shifted instances. In summary, we consider the following categories of distribution shifts: (a) background noise, (b) gender-biased training data, (c) cross-corpus generalization. While the factors of distribution shift are different in each of these cases, we consistently observe that test-time training enables robust identification  across all evaluations over both in-distribution testing samples and distributionally-shifted instances . 

\bibliographystyle{IEEEtran}
\bibliography{references.bib}

\end{document}